%% file: root.tex
\documentclass[letterpaper, 10 pt, conference]{ieeeconf}  

\IEEEoverridecommandlockouts                              

\usepackage[colorlinks,linkcolor=blue]{hyperref} 
\usepackage{mathrsfs}
\usepackage{amsfonts}
\usepackage{amsmath}
\usepackage[flushleft]{threeparttable}
\usepackage{tablefootnote}
\usepackage{multirow}
\usepackage{graphicx}
\usepackage{subfigure}
\usepackage{tikz}
\usetikzlibrary{calc,arrows,decorations.markings}
\usepackage{balance}
\usepackage{cite}
\usepackage{booktabs}
\usepackage{xpatch}
\usepackage{mathrsfs}
\usepackage{amsmath}
\usepackage{float}
\usepackage{array}    
\usepackage{tabularx}
\usepackage{titlesec}

\makeatletter
\patchcmd\@makecaption{\\}{.~}{}{\fail}
\makeatletter

\title{\LARGE \bf

HGDiffuser: Efficient Task-Oriented Grasp Generation via \\ Human-Guided Grasp Diffusion Models}

\author{Dehao Huang$^{1, 2}$, Wenlong Dong$^{1, 2}$, Chao Tang$^{1, 2}$, Hong Zhang$^{1, 2}$ \emph{Life Fellow, IEEE}
\thanks{$^{1}$Shenzhen Key Laboratory of Robotics and Computer Vision, Southern University of Science and Technology, Shenzhen, China.}%
\thanks{$^{2}$Department of Electronic and Electrical Engineering, Southern University of Science and Technology, Shenzhen, China.}%
\thanks{This work was supported in part by the Shenzhen Key Laboratory of Robotics and Computer Vision (ZDSYS20220330160557001).}
}

\begin{document}

\maketitle


\input{abstract}

\input{intro}

\input{related}

\input{approach}

\input{exp}

\input{conclusion}







\bibliographystyle{IEEEtran}
\balance
\bibliography{root}

\newpage

\end{document}

%% file: abstract.tex
\begin{abstract}
Task-oriented grasping (TOG) is essential for robots to perform manipulation tasks, requiring grasps that are both stable and compliant with task-specific constraints. Humans naturally grasp objects in a task-oriented manner to facilitate subsequent manipulation tasks. By leveraging human grasp demonstrations, current methods can generate high-quality robotic parallel-jaw task-oriented grasps for diverse objects and tasks. However, they still encounter challenges in maintaining grasp stability and sampling efficiency. These methods typically rely on a two-stage process: first performing exhaustive task-agnostic grasp sampling in the 6-DoF space, then applying demonstration-induced constraints (e.g., contact regions and wrist orientations) to filter candidates. This leads to inefficiency and potential failure due to the vast sampling space. To address this, we propose the Human-guided Grasp Diffuser (HGDiffuser), a diffusion-based framework that integrates these constraints into a guided sampling process. Through this approach, HGDiffuser directly generates 6-DoF task-oriented grasps in a single stage, eliminating exhaustive task-agnostic sampling. Furthermore, by incorporating Diffusion Transformer (DiT) blocks as the feature backbone, HGDiffuser improves grasp generation quality compared to MLP-based methods. Experimental results demonstrate that our approach significantly improves the efficiency of task-oriented grasp generation, enabling more effective transfer of human grasping strategies to robotic systems. To access the source code and supplementary videos, visit \url{https://sites.google.com/view/hgdiffuser}.
\end{abstract}

\thispagestyle{empty}
\pagestyle{empty}

%% file: intro.tex
\section{INTRODUCTION}
\begin{figure}[t]
  \centering
  \begin{tikzpicture}[inner sep = 0pt, outer sep = 0pt]
    \node[anchor=south west] (fnC) at (0in,0in)
      {\includegraphics[height=4.00in,clip=true,trim=0.18in 0.16in 0.18in 0.1in]{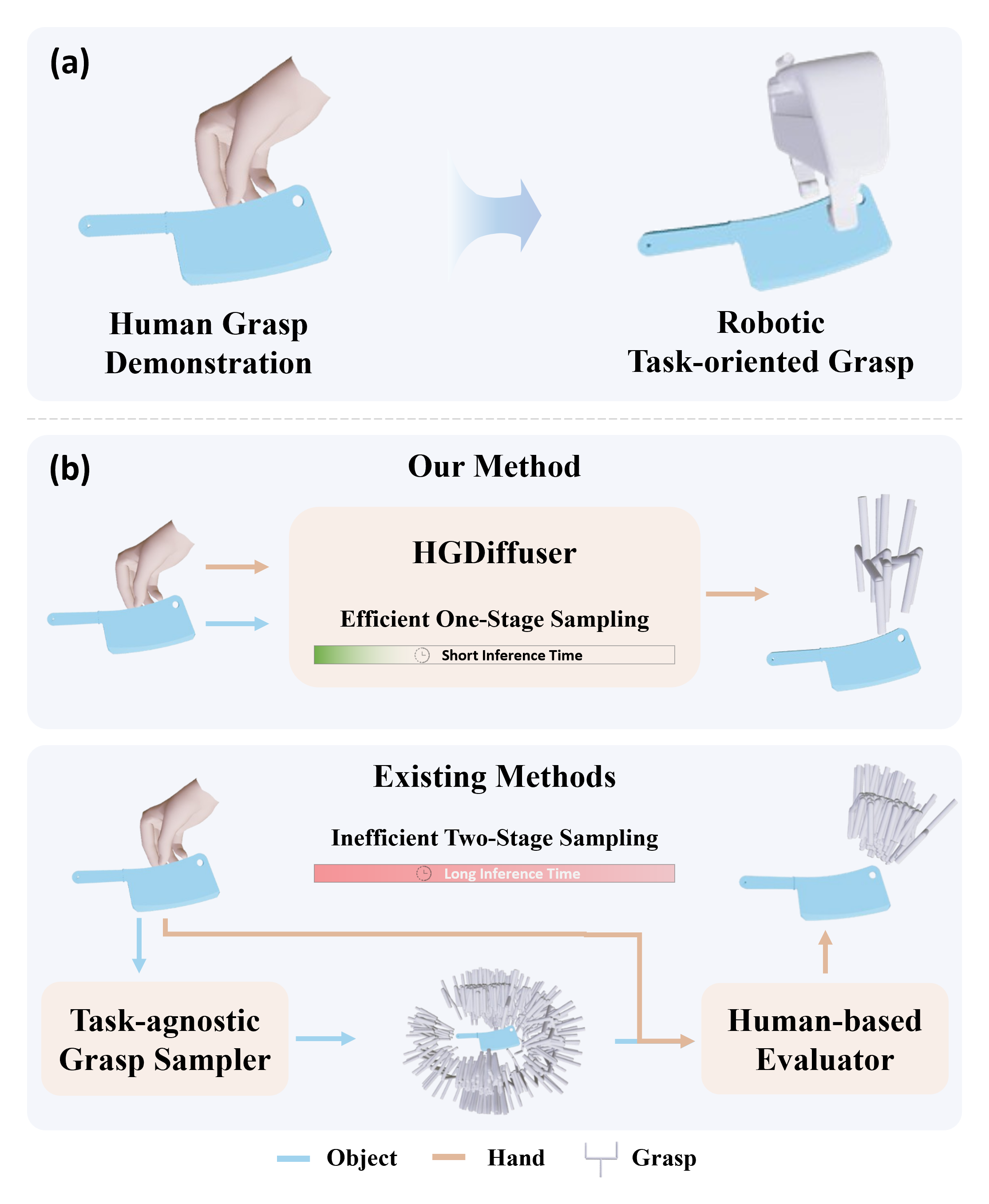}};
  \end{tikzpicture}
    \vspace*{-0.1in}
  \caption{(a) Demonstration-based methods, which generate robotic 6-DoF parallel-jaw task-oriented grasps by leveraging human demonstrations. (b) Comparison of existing two-stage methods and our single-stage method. Unlike two-stage methods, which require extensive sampling followed by filtering to generate grasps, our method directly generates grasps with minimal sampling, making it more efficient.}
  \label{fig:main}
  \vspace*{-0.30in}
\end{figure}

Task-oriented grasping (TOG) refers to grasping objects in a manner that is aligned with the intended task, which is the first and crucial step for robots to perform manipulation tasks \cite{detry2017task}. For instance, when handing over a kitchen knife to a human, the blade should be grasped perpendicular to the handle to ensure a safe and efficient handover. Many objects in daily life are designed with human convenience in mind, so human task-oriented grasping inherently includes the skills necessary for manipulating the objects, such as maintaining stability and avoiding collisions with the environment during manipulation. Consequently, previous methods have proposed to utilize human grasp demonstrations as training data or reference templates for robotic task-oriented grasp generation. In this work, we specify an important form of this problem and focus on transferring human grasps from demonstrations to robotic 6-DoF parallel-jaw task-oriented grasps due to the popularity of this type of robot end-effector, as illustrated in Figure \ref{fig:main}(a).

To transfer human grasps to robotic 6-DoF parallel-jaw task-oriented grasps, recent works \cite{dong2024rtagrasp, ju2024robo, heppert2024ditto, cai2024visual} have proposed a two-stage approach for task-oriented grasping. First, task-agnostic grasp sampling generates stable parallel-jaw grasp candidates. Subsequently, explicit task-oriented constraints derived from human demonstrations are applied to filter these candidates, ensuring both stability and task-specific requirements are met. This two-stage approach, leveraging a diverse set of stable grasp candidates, often succeeds in identifying high-quality task-oriented grasps. However, in the vast 6-DoF grasp sampling space, task-agnostic samplers require extensive sampling to generate sufficiently diverse candidates that satisfy both stability and task-oriented requirements, leading to inefficiency and a potential risk of failure.

To address this challenge, we propose the Human-guided Grasp Diffuser (HGDiffuser), a diffusion-based framework that leverages human task-oriented grasp demonstrations to directly generate 6-DoF parallel-jaw task-oriented grasps in a single sampling stage. Unlike conventional two-stage methods requiring extensive task-agnostic sampling and subsequent constraint-based filtering, HGDiffuser integrates explicit task-oriented constraints directly into the sampling process, significantly enhancing efficiency, as shown in Figure~\ref{fig:main}(b). Specifically, HGDiffuser utilizes the inherent guided sampling mechanism of diffusion models \cite{wu2024afforddp, yan2024m2diffuser, dhariwalDiffusionModelsBeat2021a}, to incorporate explicit task-oriented constraints extracted from human demonstrations—including hand-object contact regions and wrist orientations—as an additional score function to guide grasp sampling toward stable, task-compliant solutions within the 6-DoF parallel-jaw grasp manifold. By eliminating the need for exhaustive random sampling, HGDiffuser enhances efficiency and enables the direct and effective transfer of human task-oriented grasping strategies to robotic parallel-jaw grippers. 
Furthermore, building upon recent advancements in diffusion models, we implement Diffusion Transformer (DiT) blocks \cite{peebles2023scalable} as the feature backbone of HGDiffuser. This architectural choice leverages the attention mechanism inherent in transformers, enabling more effective feature fusion compared to traditional UNet-based or MLP-based backbones. The incorporation of DiT blocks improves the feature fusion of HGDiffuser. 
The experimental results demonstrate that HGDiffuer achieves a remarkable 81.26\% reduction (from 1.019s to 0.191s) in inference time compared to the state-of-the-art (SOTA) two-stage method while maintaining competitive task-oriented grasp generation quality.

In summary, the contributions of this paper are outlined as follows.
\begin{itemize}
    \item We present HGDiffuser, a novel diffusion-based framework that leverages the guided sampling mechanism to incorporate explicit task-oriented grasp constraints derived from human demonstrations. This approach eliminates the exhaustive task-agnostic sampling of conventional methods, significantly improving efficiency while maintaining grasp quality.
    \item We propose the integration of Diffusion Transformer (DiT) blocks as the core architectural component of HGDiffuser. The attention mechanism in DiT blocks enables superior feature fusion, enhancing the generated grasps' quality.
\end{itemize}

%% file: related.tex
\section{Related Work} \label{related_work}

Recent research on task-oriented grasp generation for parallel-jaw grippers can be categorized into three types based on the nature of the required data. Below, we provide a detailed discussion of each category.

\vspace{0.15\baselineskip}
\noindent\textbf{Methods based on human demonstration data.} This approach \cite{dong2024rtagrasp, ju2024robo, cai2024visual, patten2020imitating, kokic2020learning, heppert2024ditto} leverages inexpensive human grasp demonstrations to generate task-oriented grasps. Human demonstrations may include static images \cite{patten2020imitating}, videos of identical objects \cite{DemoGraspFewshotLearningwang2021, heppert2024ditto}, or videos of similar objects \cite{dong2024rtagrasp, ju2024robo} for novel object-task pairs.

A core component of these methods is the transformation module that converts human grasps into parallel-jaw grasps. Early end-to-end solutions \cite{DemoGraspFewshotLearningwang2021, patten2020imitating} directly map human grasps using manual rules or MLP networks trained on small datasets. For instance, DemoGrasp \cite{DemoGraspFewshotLearningwang2021} assumes that humans grasp objects in a fixed manner, using the midpoint between the thumb and the index finger as the parallel-jaw grasp point and the wrist orientation as the grasp direction. Patten et al. \cite{patten2020imitating} annotated a small-scale dataset of human grasps and corresponding parallel-jaw grasps, to learn a mapping using an MLP network. However, these struggle with human grasp diversity \cite{feix2015grasp} and complex mapping relationships, often yielding unstable grasps. 

Recent two-stage methods \cite{dong2024rtagrasp, cai2024visual, heppert2024ditto, ju2024robo} initially generate grasp candidates through a task-agnostic parallel-jaw grasp sampler \cite{sundermeyer2021contact}, then apply task-oriented constraints derived from human grasp demonstrations to select optimal grasps. While Robo-ABC \cite{ju2024robo} and FUNCTO \cite{tang2025functo} use region constraints, RTAGrasp \cite{dong2024rtagrasp} and DITTO \cite{heppert2024ditto} combine region and orientation constraints. This two-stage approach, leveraging a diverse set of stable grasp candidates, often identifies high-quality task-oriented grasps that satisfy both stability and task-oriented requirements. However, in the vast 6-DoF parallel-jaw grasp space, task-agnostic samplers require extensive sampling to generate sufficiently diverse candidates that satisfy both stability and task-oriented requirements, leading to inefficiency. In contrast, Our HGDiffuser addresses this by leveraging the inherent guided sampling mechanism of diffusion models to incorporate task-oriented constraints into a diffusion-based parallel-jaw task-agnostic sampler directly. This innovative single-stage approach significantly improves efficiency while reducing the risk of sampling failure. 

\vspace{0.15\baselineskip}
\noindent\textbf{Methods based on manually annotated task-oriented grasp data.} The high cost of manual annotation and generalization limitations remain key challenges for these approaches. Murali et al. \cite{murali2021same} introduce TaskGrasp, a manually curated large-scale TOG dataset, alongside GCNGrasp, which improves novel object generalization using semantic knowledge from pre-built knowledge graphs. Later works like GraspGPT \cite{tang2023graspgpt, tang2025foundationgrasp} extend capabilities by integrating open-ended semantic/geometric knowledge through LLM/VLM interactions, enabling generalization to novel object-task categories beyond training data. However, these methods remain constrained to objects/tasks resembling those in costly manual annotated datasets \cite{tang2025foundationgrasp}.

\noindent\textbf{Methods based on large-scale internet data.} Benefiting from Vision-Language Models (VLMs) pre-trained on massive web data, recent studies \cite{mirjalili2023lan, li2024seggrasp, jin2024reasoning} have explored zero-shot task-oriented grasp generation capabilities. However, these methods face two fundamental limitations: (1) the absence of fine-grained object understanding data essential for task-oriented grasping in pre-training datasets, and (2) their capability being limited to predicting task-related grasp regions. These constraints jointly lead to suboptimal performance.



%% file: approach.tex

\section{Problem Formulation}\label{problem}

\begin{figure}[t]
  \centering
  \vspace*{-0.1in}
  \begin{tikzpicture}[inner sep = 0pt, outer sep = 0pt]
    \node[anchor=south west] (fnC) at (0in,0in)
      {\includegraphics[height=2.40in,clip=true,trim=0in 0in 0in 0in]{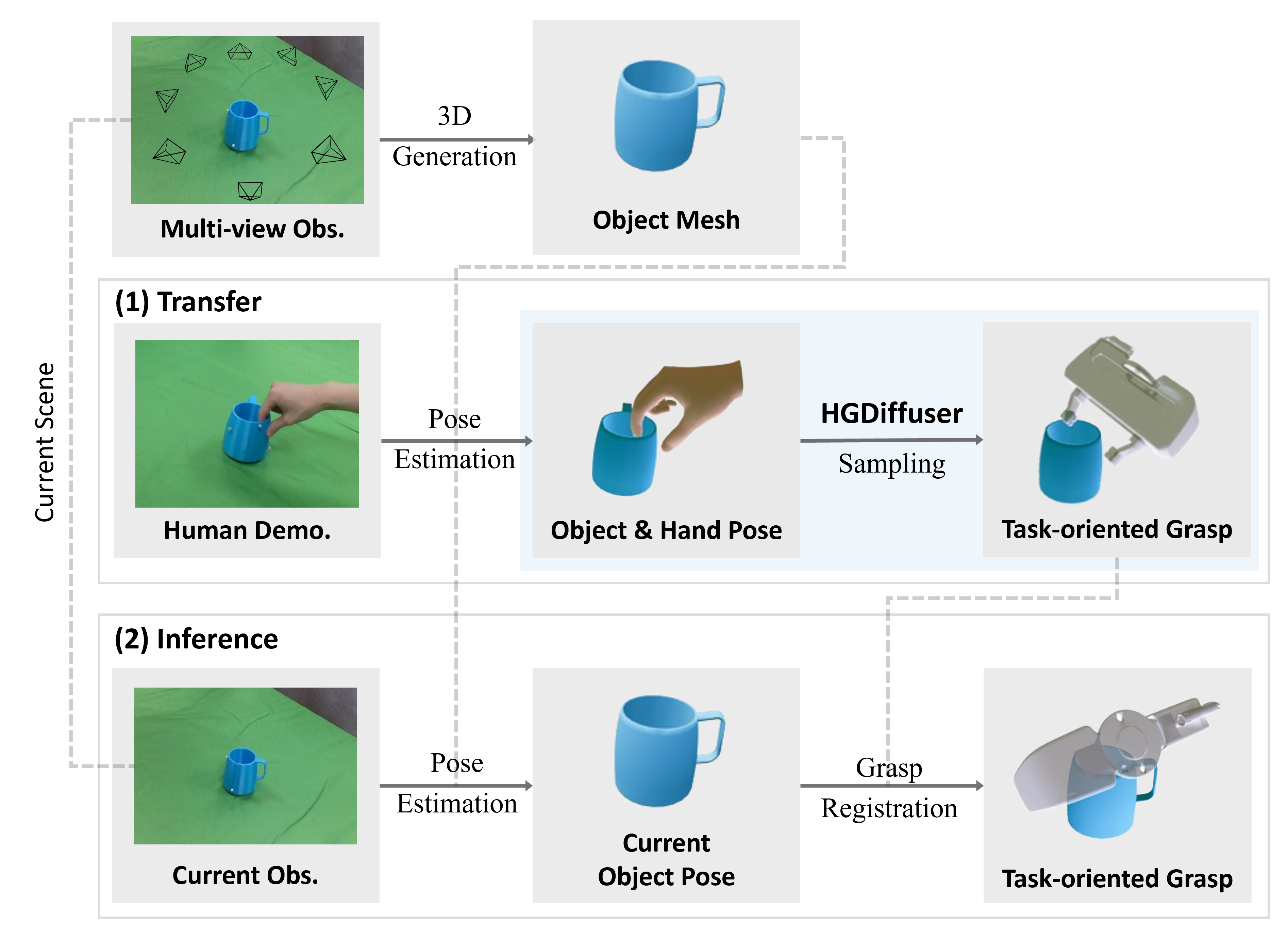}};
  \end{tikzpicture}
    \vspace*{-0.1in}
  \caption{Overview of our task-oriented grasping system. The task demonstrated is to handover a cup.}
  \label{fig:system}
  \vspace*{-0.20in}
\end{figure} 

Given a single-view RGB-D image of a human demonstration $I_{demo}$, which shows a person naturally grasping a target object $o$ in a task-oriented manner, our goal is to reproduce the same task-oriented grasp on the same target object $o$ using a parallel-jaw robotic gripper. The demonstration image $\mathbf{I}_{demo}$ can be captured by the robot’s camera.

Figure \ref{fig:system} illustrates an overview of our task-oriented grasping system. The system comprises two phases: a transfer phase and an inference phase. Before these phases, the robot performs multi-view observations of the target object $o$ and employs a 3D generation foundation model \cite{muRobotwinDualarmRobot2024a, chenG3FlowGenerative3D2024b} to reconstruct the object’s mesh $\mathbf{M}_{o}$, as shown in the upper part of Figure \ref{fig:system}.

In the transfer phase, we use $\mathbf{I}_{demo}$ and $\mathbf{M}_{o}$ as inputs, employing vision foundation models for object pose estimation \cite{wen2024foundationpose} and hand pose estimation \cite{pavlakosReconstructingHands3d2024a}. This process yields the object point cloud $\mathbf{X}_o \in \mathbb{R}^{N \times 3}$ under the current object pose and the MANO parametric model \cite{romero2022embodied} representing the human grasp $\mathbf{X}_h = \{\theta, \beta\}$, where $\theta \in \mathbb{R}^{48}$ and $\beta \in \mathbb{R}^{10}$. The task-oriented human grasp is then transformed to generate the corresponding parallel-jaw gripper grasp $\mathbf{H}$. During the inference phase, we estimate the current pose of target object $o$ and register the transformed grasp $\mathbf{H}$.

It is crucial to emphasize that our core contribution is the development of HGDiffuser for the transformation module, while other system components are implemented using established techniques.

\begin{figure*}[t]
  \centering
  \vspace*{-0.2in}
  \begin{tikzpicture}[inner sep = 0pt, outer sep = 0pt]
    \node[anchor=south west] (fnC) at (0in,0in)
      {\includegraphics[height=3.48in,clip=true,trim=0.20in 0.2in 0.20in 0.2in]{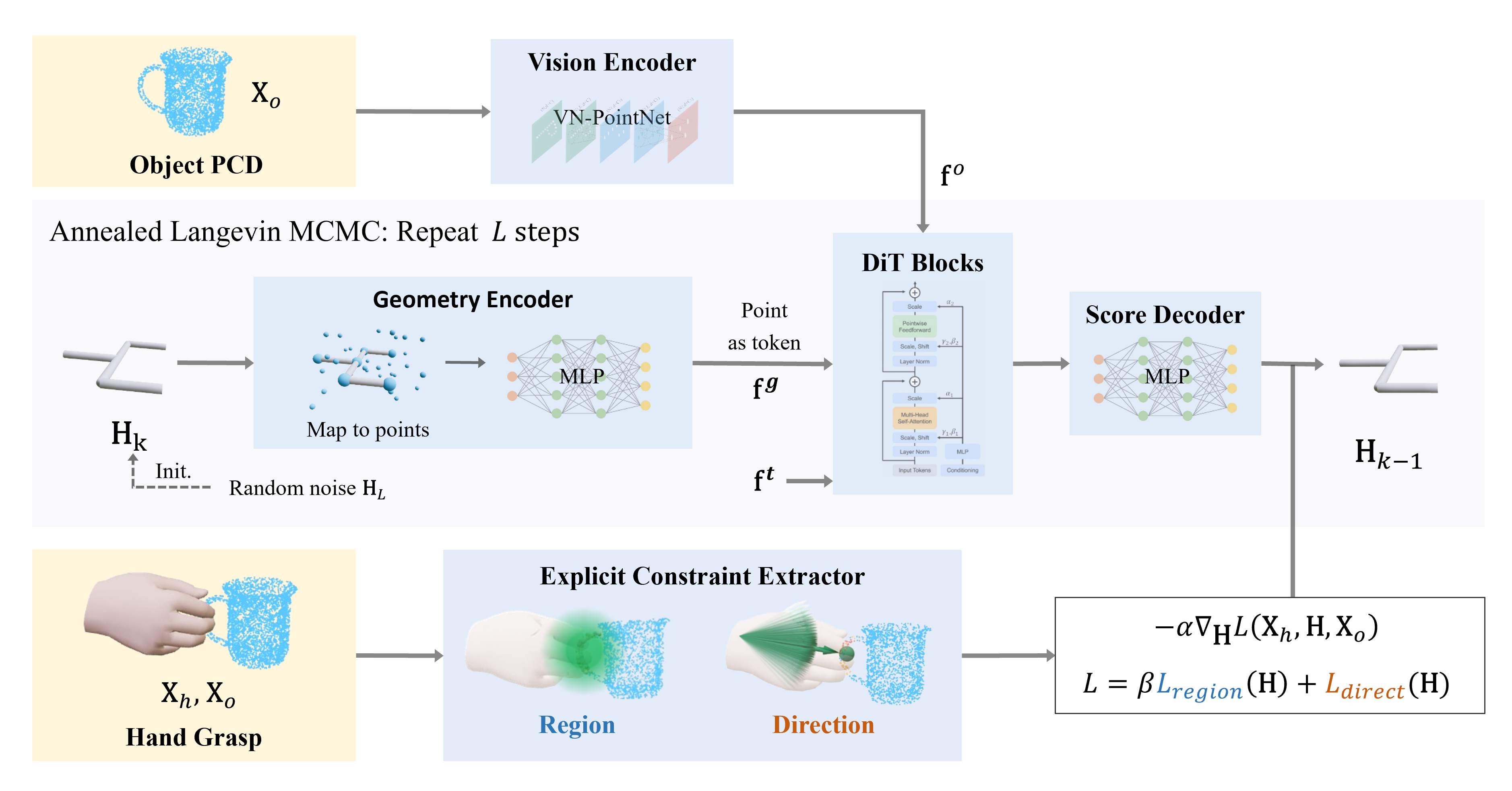}};
  \end{tikzpicture}
    \vspace*{-0.1in}
  \caption{An overview of HGDiffuser. The grasp generation employs annealed Langevin MCMC sampling with $T$ steps. The input object point cloud $\mathbf{X}_{o}$ is encoded into feature $\mathbf{f}^{o}$ via vision encoder, while current grasp $\mathbf{H}_{t}$ is processed into $\mathbf{f}^{g}$ via geometry encoder. These features, along with step feature $\mathbf{f}^{t}$ from sinusoidal encoding, serve as inputs to the DiT-based backbone. The fused features are decoded to produce a noise conditional score. For the input human grasp $\mathbf{X}_{h}$, explicit task-oriented constraints are extracted to construct a loss function guiding the sampling process. The noise conditional score, combined with the loss function, updates grasp $\mathbf{H}_{k}$ to $\mathbf{H}_{k-1}$, iterating $L$ times to output final grasp $\mathbf{H}_{0}$.}
  \label{fig:pipeline}
  \vspace*{-0.25in}
\end{figure*} 

\section{HGDiffuser}\label{method}
An overview of the proposed HGDiffuser framework is shown in Figure \ref{fig:pipeline}. Given an object point cloud $\mathbf{X}_o$ and a human grasp $\mathbf{X}_h$, HGDiffuser infers the corresponding task-oriented grasp $\mathbf{H}$ for a parallel-jaw gripper. Formally, HGDiffuser learns a conditional distribution $\rho(\mathbf{H} \mid \mathbf{X}_o, \mathbf{X}_h)$, where $\mathbf{H} \in SE(3)$ represents a valid task-oriented grasp that aligns with the human demonstration.

While prior research \cite{urain2023se, mousavian20196, barad2024graspldm} has explored VAE and diffusion models for task-agnostic grasp generation (i.e., learning $\rho(\mathbf{H} \mid \mathbf{X}_o)$), they rely on extensive simulated datasets for generalization. However, existing task-oriented grasping datasets lack the scale needed for similar end-to-end training. To address this problem, our approach leverages the guidance mechanism of diffusion models \cite{dhariwalDiffusionModelsBeat2021a}. We train a diffusion-based generative model on task-agnostic data to learn $\rho(\mathbf{H} \mid \mathbf{X}_o)$. During inference, we employ the guided sampling mechanism to incorporate the explicit task-oriented constraints $\rho(\mathbf{X}_h \mid \mathbf{H})$ derived from human grasp demonstrations, effectively sampling from $\rho(\mathbf{H} \mid \mathbf{X}_o, \mathbf{X}_h)$. This design allows us to generate task-oriented grasps that are stable and aligned with human demonstrations without requiring large-scale task-oriented training data.


\subsection{DiT-based Diffusion Model for $\rho(\mathbf{H} \mid \mathbf{X}_o)$ }
For the first part of HGDiffuser, we build on the prior diffusion-based task-agnostic grasp sampler \cite{urain2023se} and introduce the DiT blocks as the feature backbone.

\vspace{0.1\baselineskip}
\noindent\textbf{Training procedure.} The Denoising Score Matching (DSM) \cite{hyvarinen2005estimation} is employed as the training procedure. Given a grasp dataset with $\{\mathbf{H}, \mathbf{X}_o\}$, the goal is to use a Noise Conditional Score Network (NCSN) $\mathbf{s}_\theta$ to learn $\rho(\mathbf{H} \mid \mathbf{X}_o)$. The NCSN $\mathbf{s}_\theta(\mathbf{H}, k, \mathbf{X}_o)$ is trained to estimate $\nabla_{\mathbf{H}} \log \rho_{\sigma_k}(\mathbf{H} \mid \mathbf{X}_o)$, where $k \in \{0, \ldots, L-1\}$ denotes a noise scale among levels $\sigma_k$. The training objective minimizes the following loss function $\mathcal{L}_{\mathrm{dsm}}$, using a score matching method \cite{song2020improved}:

\vspace*{-0.9\baselineskip}
\begin{small} 
    \begin{align*}
    \mathcal{L}_{\mathrm{dsm}} = \sum_{k=1}^L \mathbb{E}_{\rho_{\sigma_i}(\mathbf{H} \mid \mathbf{X}_o)}\bigl[&\left\|\nabla_{\mathbf{H}} \log \rho_{\sigma_k}(\mathbf{H} \mid \mathbf{X}_o)\right. \\
    &\left.-\mathbf{s}_\theta(\mathbf{H}, k, \mathbf{X}_o)\right\|_2^2\bigr]
    \end{align*}
\end{small}
\vspace*{-1.3\baselineskip}

\vspace{0.1\baselineskip}
\noindent\textbf{Inference procedure.} The trained NSCN generates samples based on annealed Langevin Markov chain Monte Carlo (MCMC) \cite{song2020improved}. A sample $\mathbf{H}_{L}$ is drawn from $\mathcal{N}(\mathbf{0}, \mathbf{I})$, followed by $L$-step iterations (from $k=L-1$ to $k=0$):

\vspace*{-0.9\baselineskip}
\begin{small}
\begin{align*}
\mathbf{H}_{k-1} &= \mathbf{H}_k + \epsilon_k \nabla_{\mathbf{H}} \log \rho_{\sigma_k}(\mathbf{H} \mid \mathbf{X}_o) + \sqrt{2 \epsilon} \mathbf{z} \\
                 &= \mathbf{H}_k + \epsilon_k \mathbf{s}_\theta(\mathbf{H}, k, \mathbf{X}_o) + \sqrt{2 \epsilon} \mathbf{z}, \quad \mathbf{z} \sim \mathcal{N}(\boldsymbol{0}, \boldsymbol{I})
\end{align*}
\end{small}
\vspace*{-0.9\baselineskip}

\noindent where $\epsilon_k$ is a step-dependent coefficient that decreases as $k$ decreases. The inference process repeats the central part of Figure \ref{fig:pipeline} for $L$ steps.

\vspace{0.1\baselineskip}
\noindent\textbf{Network architecture.} The NSCN $\mathbf{s}_\theta(\mathbf{H}, k, \mathbf{X}_o)$ takes as input the object point cloud $\mathbf{X}_o$, the current noise step $k$, and the current grasp $\mathbf{H}_k$, and outputs the score $\mathbf{H}_s \in SE(3)$. The object point cloud $X_o$ is encoded using VN-PointNet \cite{deng2021vector}, a SO(3)-equivariant point cloud feature encoder, producing the feature vector $\mathbf{f}^{o} \in \mathbb{R}^{d}$, where $d$ denotes the descriptor dimension. For the current grasp $\mathbf{H}_k \in SE(3)$, a predefined gripper points mapper first transforms it into gripper points $\mathbf{X}_g \in \mathbb{R}^{g \times 3}$, as illustrated in Figure \ref{fig:pipeline}, with $g$ being the predefined number of points. Encoding the grasp via gripper points rather than directly from $SE(3)$ allows for more effective integration with the object point cloud features, as both are represented in Cartesian space. The gripper points $\mathbf{X}_g$ are then processed by an MLP to generate the feature vector $\mathbf{f}^{g} \in \mathbb{R}^{g \times d}$. Finally, the noise step $k$ is encoded using transformer sinusoidal position embedding \cite{peebles2023scalable}, resulting in the feature vector $\mathbf{f}^{t} \in \mathbb{R}^{d}$.

To effectively fuse the features $\{\mathbf{f}^{g}, \mathbf{f}^{o}, \mathbf{f}^{t}\}$, we introduce a DiT-based feature backbone. At its core lies a transformer block with an attention mechanism, which has been extensively validated to achieve superior feature fusion capabilities across most tasks compared to other alternatives. Our DiT-based feature backbone comprises $D$ DiT blocks \cite{peebles2023scalable} connected in series, where $D$ is a configurable parameter adjustable based on data volume and model size. The DiT block, a variant of standard transformer block, takes input tokens and condition features as inputs and outputs fused feature tokens. It incorporates the condition feature input into the block's adaptive layer normalization, a method demonstrated to be more efficient and effective than direct fusion using cross-attention.

Since our diffusion model learns the distribution of grasps $\mathbf{H}$, denoted as $\rho(\mathbf{H} \mid \mathbf{X}_o)$, the gripper points feature $\mathbf{f}^{g}$ serves as the input tokens, while the fused object point cloud feature $\mathbf{f}^{o}$ and noise step feature $\mathbf{f}^{t}$ are utilized as the condition feature input. For the gripper points feature $\mathbf{f}^{g} \in \mathbb{R}^{g \times d}$, we propose a method inspired by image transformers, where each image patch is treated as a token. Existing methods for handling point cloud features as transformer input tokens \cite{wu2024point} typically rely on advanced point serialization strategies and serialized attention mechanisms to address the unordered nature and large quantity of point cloud data. In our approach, however, the predefined gripper points mapper directly converts grasp $\mathbf{H}$ into a serialized set of gripper points, analogous to the serialization of pixels in images. Consequently, we treat each gripper point feature as an individual input token. Following established practices for processing noise step features, we directly sum the object point cloud feature $\mathbf{f}^{o}$ and the noise step feature $\mathbf{f}^{t}$ to form the condition feature input $\mathbf{f}^{c} = \mathbf{f}^{t} + \mathbf{f}^{o}$.

\subsection{Guidance-based Inference for $\rho(\mathbf{H} \mid \mathbf{X}_o, \mathbf{X}_h)$}
After obtaining the diffusion model that has learned $\rho(\mathbf{H} \mid \mathbf{X}_o)$, representing task-agnostic grasp generation, we leverage the guided sampling mechanism inherent in diffusion models to incorporate explicit task-oriented constraints derived from human grasp demonstration $\mathbf{X}_h$. This enables the diffusion model to sample from the distribution $\rho(\mathbf{H} \mid \mathbf{X}_o, \mathbf{X}_h)$, corresponding to the desired task-oriented grasp generation.

\vspace{0.1\baselineskip}
\noindent\textbf{Inference procedure.} To sample from the distribution $\rho(\mathbf{H} \mid \mathbf{X}_o, \mathbf{X}_h)$, we follow the approach of the aforementioned diffusion model by transforming the problem into estimating the score function $\nabla_{\mathbf{H}} \log \rho(\mathbf{H} \mid \mathbf{X}_o, \mathbf{X}_h)$. 
Using Bayes' theorem, the score function of the conditional distribution decomposes into the sum of the score functions of the prior distribution and the likelihood distribution:

\vspace*{-0.9\baselineskip}
\begin{small}
\begin{align*}
\nabla_{\mathbf{H}} \log \rho(\mathbf{H} & \mid \mathbf{X}_o, \mathbf{X}_h) = \\
&\nabla_{\mathbf{H}} \log \rho(\mathbf{H} \mid \mathbf{X}_o) + \nabla_{\mathbf{H}} \log \rho(\mathbf{X}_h \mid \mathbf{H}, \mathbf{X}_o)
\vspace*{-0.9\baselineskip}
\end{align*}
\end{small}

\begin{figure*}[t]
  \centering
  \vspace*{-0.3in}
  \begin{tikzpicture}[inner sep = 0pt, outer sep = 0pt]
    \node[anchor=south west] (fnC) at (0in,0in)
      {\includegraphics[height=2.68in,clip=true,trim=0.05in 0.1in 0in 0in]{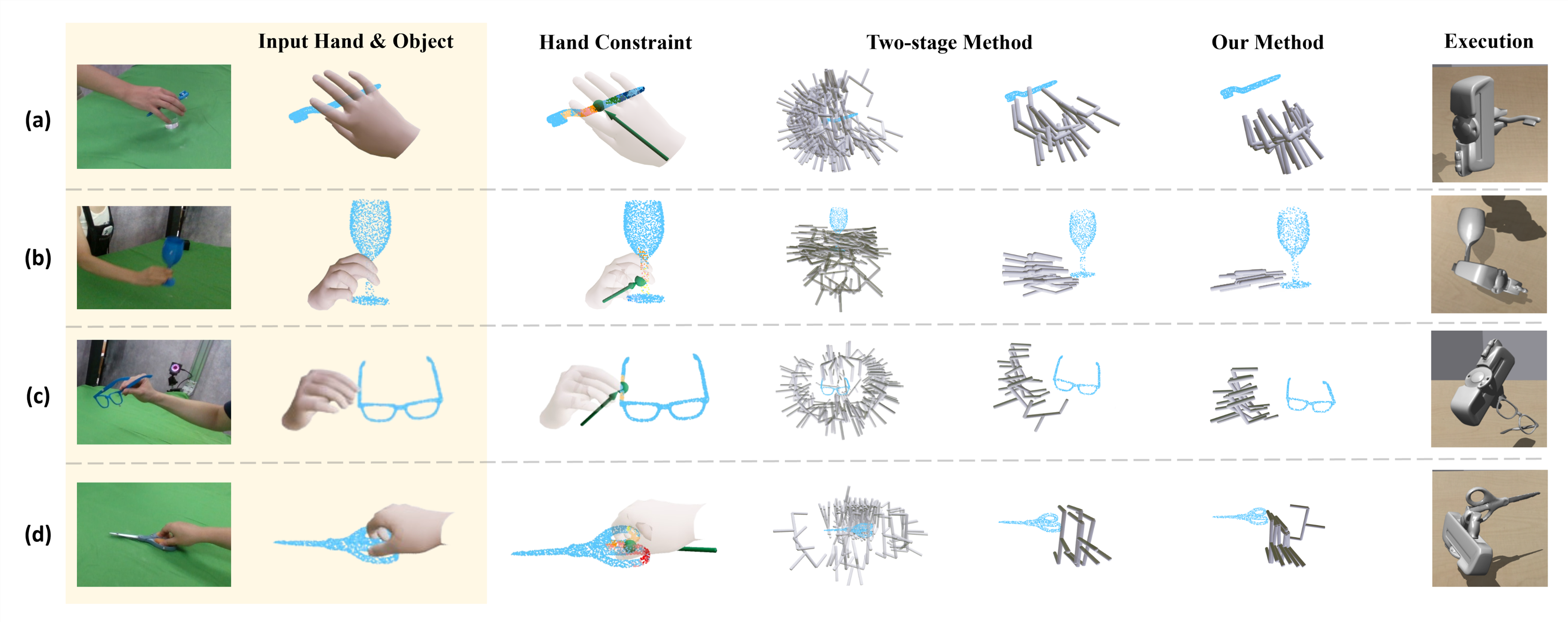}};
  \end{tikzpicture}
    \vspace*{-0.12in}
  \caption{Qualitative results of our method and Ours-TS method. The object categories and tasks are as follows: (a) toothbrush and brushing, (b) wine glass and pouring, (c) eyeglasses and handing over, (d) scissors and using. More results are provided in the supplementary material.}
  \label{fig:exp}
  \vspace*{-0.24in}
\end{figure*}

Here, the prior distribution component $\nabla_{\mathbf{H}} \log \rho(\mathbf{H} \mid \mathbf{X}_o)$ corresponds to our trained NSCN $\mathbf{s}_\theta(\mathbf{H}, k, \mathbf{X}_o)$. The likelihood distribution component $\nabla_{\mathbf{H}} \log \rho(\mathbf{X}_h \mid \mathbf{H}, \mathbf{X}_o)$ represents guidance derived from human grasp demonstration. In the context of class-conditional image generation tasks, this term is known as Classifier Guidance \cite{dhariwalDiffusionModelsBeat2021a}, typically estimated using a pre-trained image classifier. The image classifier's understanding of image categories guides unconditionally generative models in generating images of specific categories. In our task, we leverage the task-relevant information in human grasps to guide the generation of task-oriented parallel-jaw grasps. Specifically, we introduce a non-network-based, differentiable loss function $L(\mathbf{X}_h, \mathbf{H}, \mathbf{X}_o)$, which evaluates the probability of a parallel-jaw grasp $\mathbf{H}$ meeting task-oriented requirements based on the human grasp demonstration $\mathbf{X}_h$. The lower the loss, the higher the corresponding probability. The estimated score function $\nabla_{\mathbf{H}} \log \rho(\mathbf{H} \mid \mathbf{X}_o, \mathbf{X}_h)$ is expressed as:

\vspace*{-0.9\baselineskip}
\begin{small} 
\begin{align*}
\nabla_{\mathbf{H}} \log \rho(\mathbf{H} \mid \mathbf{X}_o, \mathbf{X}_h) &= \mathbf{s}_\theta(\mathbf{H}, k, \mathbf{X}_o) - \alpha\nabla_{\mathbf{H}} L(\mathbf{X}_h, \mathbf{H}, \mathbf{X}_o)
\vspace*{-0.9\baselineskip}
\end{align*}
\end{small}

\noindent where $\alpha$ is a scaling parameter. Sampling from the distribution $\rho(\mathbf{H} \mid \mathbf{X}_o, \mathbf{X}_h)$ using the annealed Langevin MCMC method follows the equation:

\vspace*{-0.9\baselineskip}
\begin{small} 
\begin{align*}
\mathbf{H}_{k-1} &= \mathbf{H}_k + \epsilon_k [\mathbf{s}_\theta(\mathbf{H}, k, \mathbf{X}_o)-\alpha\nabla_{\mathbf{H}} L(\mathbf{X}_h, \mathbf{H}, \mathbf{X}_o)] + \sqrt{2 \epsilon} \mathbf{z}
\vspace*{-0.9\baselineskip}
\end{align*}
\end{small}

\noindent\textbf{Explicit task-oriented constraint extraction.} As illustrated in the lower part of Figure \ref{fig:pipeline}, we extract explicit task-oriented constraints from the human grasp demonstration to compute the loss function $L(\mathbf{X}_h, \mathbf{H}, \mathbf{X}_o)$, which serves as the guidance. To satisfy the task-oriented requirements of parallel-jaw grasps, two constraints are typically essential: the task-oriented grasp region constraint and the task-oriented grasp orientation constraint \cite{tang2023graspgpt, dong2024rtagrasp}. For instance, in using a brush to clean a table, the parallel-jaw gripper should grasp the brush handle in an upright orientation away from the bristle end. This combination of grasp region and orientation ensures minimal collision with the environment (e.g., the table or debris) during the cleaning process. It reduces the force required to hold the brush.

These constraints are naturally embedded in the task-oriented human grasp, which inherently contains information about both aspects. We establish two corresponding loss functions to formalize these constraints: $L_{region}$ and $L_{direct}$. Specifically, since the task-oriented grasp region of the human hand often overlaps significantly with that of the parallel-jaw gripper, we compute all contact points \cite{yang2021cpf} from $\mathbf{X}_h$ and $\mathbf{X}_o$, then derive the center point $P_{region}$. $L_{region}$ is formulated as a RELU function based on the Euclidean distance to $P_{region}$, which remains zero when the distance is below a predefined threshold $\tau_{region}$ and increases linearly beyond it, as follows:

\vspace*{-0.9\baselineskip}
\begin{small} 
\begin{align*}
L_{\text{region}}(\mathbf{H}) = \text{RELU}\left(\left\|\operatorname{Trans}(\mathbf{H}) - P_{region}\right\|_2 - \tau_{region}\right)
\end{align*}
\end{small} 
\vspace*{-0.9\baselineskip}

Here, the $\operatorname{Trans}$ function extracts the translational component of $\mathbf{H}$, and $\tau_{region}$ is the predefined distance threshold. Regarding grasp orientation, prior research has demonstrated that human grasps often avoid collisions between the arm and the environment during task execution \cite{saito2022task, wang2024robot}. To ensure the parallel-jaw gripper's relative position resembles that of the human hand, we align its orientation with the human wrist orientation. From the human hand $\mathbf{X}_h$, we compute the corresponding wrist orientation $D_{direct} \in SO(3)$ \cite{romero2022embodied}. $L_{direct}$ is formulated as a RELU function based on the geodesic distance to $D_{direct}$, as follows:

\vspace*{-0.9\baselineskip}
\begin{small} 
\begin{align*}
L_{\text{direct}}(\mathbf{H}) = \text{RELU}\left(\operatorname{GeoDistance}\left(\operatorname{Rot}(\mathbf{H}), D_{direct}\right) -\tau_{direct}\right)
\end{align*}
\end{small} 
\vspace*{-0.9\baselineskip}

The $\operatorname{Rot}$ function extracts the rotational component of $\mathbf{H}$, $\operatorname{GeoDistance}$ computes the geodesic distance, and $\tau_{direct}$ is the predefined angular threshold. The final loss function is obtained as the weighted sum of $L_{region}$ and $L_{direct}$:

\vspace*{-0.9\baselineskip}
\begin{small} 
\begin{align*}
L = \beta L_{region} + L_{direct}
\end{align*}
\end{small} 
\vspace*{-0.9\baselineskip}

\noindent where $\beta$ is a scaling parameter.

%% file: exp.tex
\section{Experimental Results} \label{exp}
In this section, we compare HGDiffuser with two categories of existing methods for generating 6-DoF parallel-jaw task-oriented grasps from human demonstrations: (1) rule-based direct conversion methods and (2) two-stage frameworks involving candidate generation and task-constrained filtering. This comparison aims to evaluate the efficiency of HGDiffuser. We also examine the impact of DiT blocks through an ablation study. Furthermore, we assess the practical applicability of HGDiffuser through real-world experiments.

\begin{figure*}[t]
  \centering
  \vspace*{-0.15in}
  \begin{tikzpicture}[inner sep = 0pt, outer sep = 0pt]
    \node[anchor=south west] (fnC) at (0in,0in)
      {\includegraphics[height=3.58in,clip=true,trim=0.38in 0.38in 0.38in 0.48in]{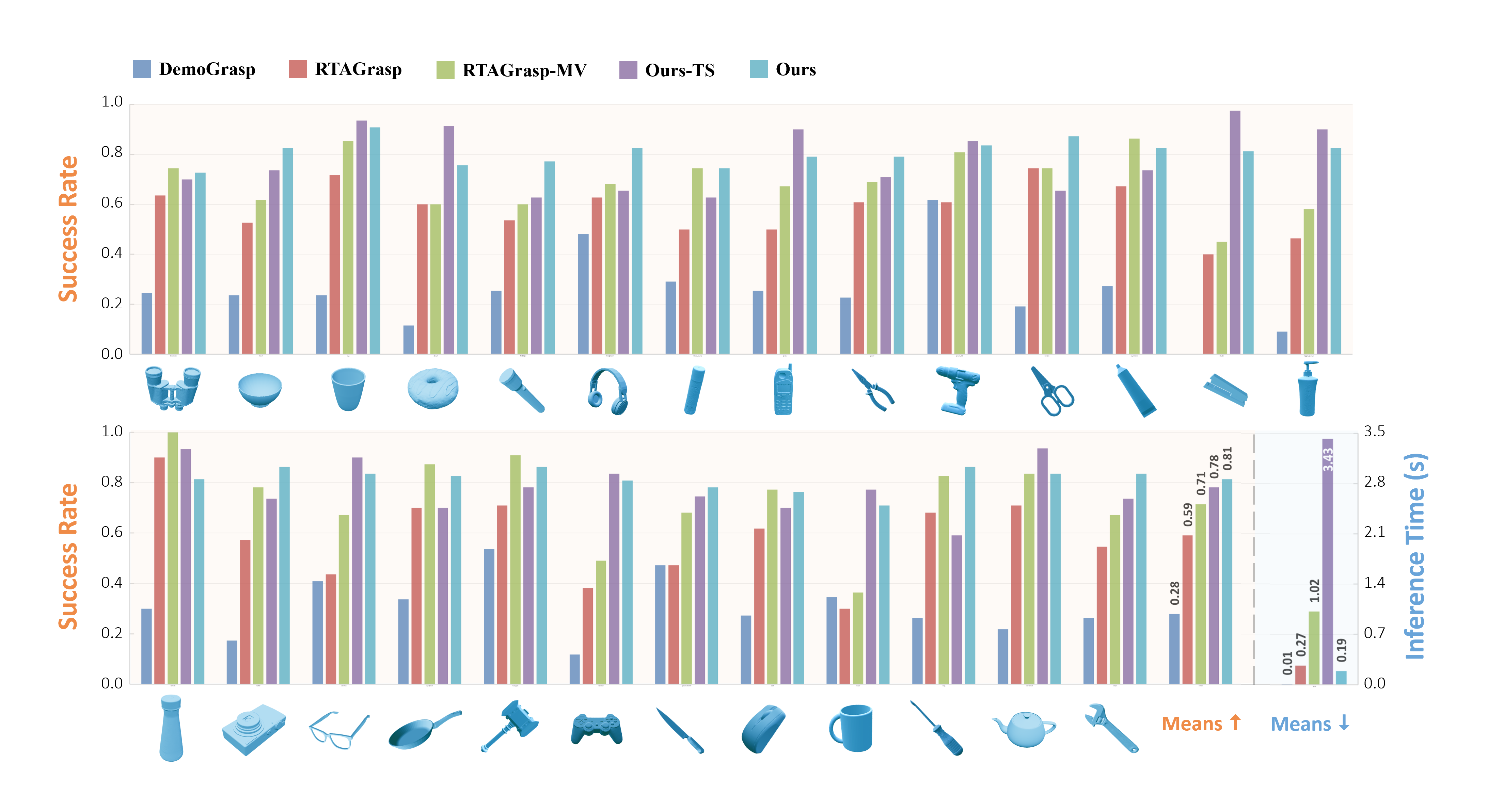}};
  \end{tikzpicture}
    \vspace*{-0.26in}
  \caption{Quantitative results. In the bottom-right section, we compare our method with baseline approaches in terms of average task-oriented grasping success rate and average inference time. The remaining sections present the average success rates across 24 object categories (out of 236 total object instances) from the dataset.}
  \label{fig:quanti_result}
  \vspace*{-0.22in}
\end{figure*} 

\subsection{Quantitative Comparison}
\noindent \textbf{Baselines} We compare HGDiffuser to the following methods: (1) DemoGrasp \cite{DemoGraspFewshotLearningwang2021}, which generates parallel-jaw task-oriented grasps based on the direct conversion of human task-oriented grasps using manually designed rules. Specifically, the grasp direction of the parallel-jaws corresponds to the wrist orientation, while the grasp point is aligned with the center of the contact region between the index finger, thumb, and the object. (2) RTAGrasp \cite{dong2024rtagrasp}, Robo-ABC \cite{ju2024robo} and DITTO \cite{heppert2024ditto}, all of which employ a similar two-stage approach, represent state-of-the-art methods for generating parallel-jaw task-oriented grasps from human demonstrations. The first stage utilizes Contact-GraspNet \cite{sundermeyer2021contact} as a task-agnostic grasp sampler to generate candidate grasps from the object’s point cloud. In the second stage, these candidates are filtered based on explicit task-oriented constraints derived from human demonstrations, mainly considering wrist orientation and hand-object contact points. In our results, we collectively refer to these three methods as RTAGrasp.

Since Contact-GraspNet operates on partially observed point clouds, these methods rely on partial observations as input. In contrast, HGDiffuser utilizes the complete object point cloud. To ensure a fair comparison, we aggregate grasps generated by Contact-GraspNet from four different viewpoints, aiming for a more comprehensive coverage of the object’s surface. This enhanced version is denoted as RTAGrasp-MultiView (RTAGrasp-MV). Additionally, we introduce a two-stage variant of our approach, which follows a method similar to the two-stage baselines but replaces Contact-GraspNet with our sampler, which does not incorporate human demonstration guidance. This variant is denoted as Ours-Two-stage (Ours-TS), enabling a more equitable comparison between our single-stage and existing two-stage methods.

\vspace{0.1\baselineskip}
\noindent \textbf{Dataset} We evaluate HGDiffuser and baselines on the OakInk dataset \cite{yang2022oakink}. Specifically, we use a subset from OakInk-Shape, selecting 340 object instances across 33 object categories, with approximately 10 instances randomly chosen per category. Each object instance includes an object mesh and a corresponding task-oriented human grasp demonstration. The object mesh is used to generate object point clouds via sampling, while the human grasp, represented using MANO representation, is directly used as input.

\vspace{0.1\baselineskip}
\noindent \textbf{Metrics} Following prior works, we evaluate the task-oriented grasping success for each object instance and compute the average success rate across all instances. The stability of task-oriented grasps is automatically evaluated using the NVIDIA Isaac Gym simulation platform \cite{makoviychuk2021isaac}, while the task relevance is manually assessed based on the corresponding human demonstration. Additionally, we evaluate the inference time to compare the efficiency of different methods.

\vspace{0.1\baselineskip}
\noindent \textbf{Implementation Details} All experiments are conducted on a desktop PC equipped with a single Nvidia RTX 3090 GPU. HGDiffuser is optimized using the Adam optimizer \cite{kingma2014adam} with a weight decay of 0.0001 and a learning rate of 0.0001. The model is trained for 500 epochs with a batch size of 32. 


\begin{table}[H]
    \centering
    \vspace*{-0.05in}
    \caption{Comparison of Different Grasp Sampling Quantities}
    \label{table:time}
    \vspace*{-0.1in}
    \renewcommand{\arraystretch}{1.15} 
    \begin{tabularx}{\columnwidth}{>{\centering\arraybackslash}p{1.2cm}>{\centering\arraybackslash}p{0.8cm}>{\centering\arraybackslash}X>{\centering\arraybackslash}X}
    \toprule
    Method & \# & Success Rate (\%) & Inference Time (s) \\ \midrule
    \multirow{4}{*}{Ours-TS} & 100  & 71.18             & \underline{0.671}              \\
                             & 200  & 75.35             & 1.352              \\
                             & 500  & 78.12             & 3.428              \\
                             & 1000 & 77.68             & 6.793              \\ \midrule
    \multirow{2}{*}{Ours}    & 1    & \underline{81.21}             & \textbf{0.191}              \\
                             & 100  & \textbf{81.38}             & 0.716              \\ \bottomrule
    \end{tabularx}
    \vspace*{-0.18in}
\end{table}

\vspace{0.2\baselineskip}
\noindent \textbf{Results} Figure~\ref{fig:quanti_result} presents the comparison results against the baseline methods. DemoGrasp achieves a success rate of only 27.85\%, primarily because only human demonstrations that conform to manually designed rules can be successfully converted into stable grasps. This limitation aligns with our observations in Section \ref{related_work}. RTAGrasp, leveraging a two-stage framework, improves the success rate to 59.06\%. However, since grasp candidates are generated from single-view point clouds, they often fail to cover the entire object, potentially missing the grasps corresponding to human demonstrations, which leads to unsuccessful attempts. RTAGrasp-MV aggregates grasps from multiple viewpoints, providing more comprehensive object coverage and improving the success rate to 71.47\%. However, this enhancement comes at the cost of increased inference time, reaching 1.019s due to the additional viewpoint processing. Our two-stage variant, Ours-TS,  achieves a success rate of 78.12\% but requires the longest inference time of 3.428s. In contrast, our HGDiffuser (Ours) integrates human demonstration guidance directly into the sampling process, effectively filtering out grasp candidates that fail to satisfy task constraints in the first stage. As a result, HGDiffuser not only outperforms both RTAGrasp-MV and Ours-TS in terms of success rate but also significantly reduces inference time. 

The number of grasp samples significantly impacts both success rate and inference time. For two-stage methods, increasing the number of first-stage grasp samples enhances the likelihood of filtering more stable, task-oriented grasps during the second stage. Table~\ref{table:time} presents the performance of Ours-TS and Ours under different grasp sampling quantities. As the number of sampled grasps increases, Our-TS success rate improves from 71.18\% to 78.12\%, but at the expense of significantly longer inference time, ranging from 0.671s to 3.428s. In contrast, Ours is almost unaffected by the number of grasp samples and can achieve the highest success rate by sampling just a single grasp. Ours-TS employs our grasp sampler without human guidance, which maintains consistent performance in 6-DoF grasp sampling compared to Ours. This result demonstrates the significant efficiency advantage of our single-stage sampling over the two-stage approach.

Figure \ref{fig:exp} presents the qualitative experimental results of Ours-TS and Ours. Combined with the inference time evaluation in Table \ref{table:time}, it can be observed that our method achieves comparable grasp generation quality to two-stage methods while significantly reducing inference time. 


\subsection{Ablation Study}
\vspace*{-0.01in}
We conduct our evaluation on the same OakInk dataset as in previous experiments. Since the introduced DiT blocks are expected to enhance the overall performance of the sampler and to ensure comparability with other existing samplers, we evaluate the performance of the sampler in generating task-agnostic grasps, considering only whether the grasp succeeds in the simulation platform without imposing task constraints. To establish a strong baseline, we include the state-of-the-art diffusion-based grasp sampler GraspLDM \cite{barad2024graspldm} in our evaluation. Additionally, we assess a variant of our method without DiT blocks (Ours w/o DiT), which corresponds to the existing approach GraspDiff \cite{urain2023se}, utilizing an MLP-based feature backbone.

\begin{table}[t]
    \centering
    \vspace*{-0.15in}
    \caption{Ablation Study on Feature Backbone}
    \label{table:ablation}
    \vspace*{-0.08in}
    \renewcommand{\arraystretch}{1.15} 
    \begin{tabularx}{\columnwidth}{>{\centering\arraybackslash}X>{\centering\arraybackslash}X>{\centering\arraybackslash}X}
    \toprule
    Method & Success Rate (\%) & Inference Time (s) \\ \midrule
    GraspLDM & 74.32 & 0.183 \\ \midrule
    Ours w/o DiT & 71.35 & 0.163 \\
    Ours & 80.65 & 0.167 \\ \bottomrule
    \end{tabularx}
    \vspace*{-0.28in}
\end{table}

Table \ref{table:ablation} presents the ablation study results. The Ours w/o DiT variant achieves an average grasp success rate of 71.35\%, which is 2.97\% lower than the state-of-the-art GraspLDM. By incorporating DiT blocks as feature backbones and designing corresponding tokenized inputs, our full model Ours attains a success rate of 80.65\%, surpassing GraspLDM by 6.33\%, while maintaining a nearly unchanged inference time. These results demonstrate that the integration of DiT blocks effectively leverages attention mechanisms to extract more informative features, thereby improving the performance of HGDiffuser.

\subsection{Real-world Experiments}
\vspace*{-0.01in}
To assess the practicality and applicability of our proposed method, we perform a quantitative evaluation through physical experiments, with the results presented in Table~\ref{table:realexp}. The videos of a subset of these experiments are provided in the Supplementary Materials. Our experimental setup utilizes a table-mounted 7-DoF Franka Research 3 arm with a Franka hand, enhanced by a wrist-mounted RealSense D435i camera. We conducted experiments on 30 object-task pairs (including 10 objects and eight tasks), with one human demonstration collected for each pair. Following the evaluation framework of \cite{tang2023graspgpt}, we measure success rates across three stages: perception, planning, and action. Our system achieves an 86.67\% success rate in the perception stage, validating the practical applicability of our method.

\begin{figure}[t]
  \centering
  \vspace*{-0.15in}
  \begin{tikzpicture}[inner sep = 0pt, outer sep = 0pt]
    \node[anchor=south west] (fnC) at (0in,0in)
      {\includegraphics[height=0.9in,clip=true,trim=0in 0.1in 0in 0.1in]{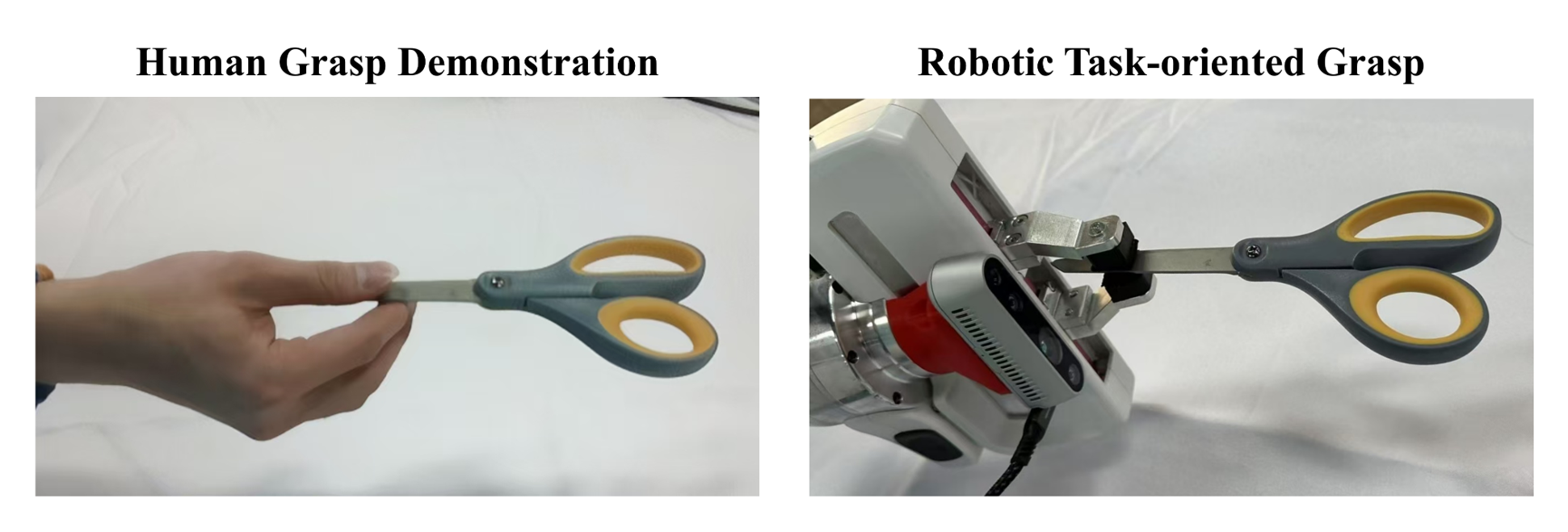}};
  \end{tikzpicture}
    \vspace*{-0.12in}
  \caption{An example of real-world experiments.}
  \label{fig:metrics}
  \vspace*{-0.08in}
\end{figure} 

\begin{table}[t]
    \centering
    \caption{Quantitative Results of Real-World Experiments}
    \label{table:realexp}
    \vspace*{-0.1in}
    \renewcommand{\arraystretch}{1.15} 
    \begin{tabularx}{\columnwidth}{>{\centering\arraybackslash}X>{\centering\arraybackslash}X>{\centering\arraybackslash}X>{\centering\arraybackslash}X}
    \toprule
    Stage & Perception & Planning & Action \\ \midrule
    Success Rate & 26 / 30 & 22 / 30 & 20 / 30 \\ \midrule
    \end{tabularx}
    \vspace*{-0.32in}
\end{table}

Physical experiments identify two perception challenges hindering task-oriented grasping in our method. First, the task-agnostic grasp sampler, trained on a large-scale dataset \cite{eppner2021acronym}, often fails to generate stable grasps for specific object regions, such as the headband of headphones or the handles of scissors and teapots. Even with human-guided sampling, the resulting grasps remain unstable. Second, discrepancies between the object's pose during grasping and demonstration necessitate pose estimation \cite{wen2024foundationpose}. However, inaccurate reconstructed meshes or partial occlusion lead to pose estimation errors, causing imprecise grasps—a common issue in demonstration-based methods.

%% file: conclusion.tex
\section{Conclusion} \label{conclusion}
\vspace*{-0.01in}
In this work, we propose HGDiffuser, a diffusion-based framework that leverages human grasp demonstrations to generate robotic 6-DoF parallel-jaw task-oriented grasps. The human grasp demonstrations are processed to create explicit task-oriented constraints, which are then used to guide the sampling of a pre-trained task-agnostic diffusion model. Compared to existing two-stage methods, HGDiffuser eliminates the need for extensive sampling in the vast task-agnostic grasp space, resulting in significantly higher efficiency and comparable or higher accuracy than the approaches. Evaluation on the OakInk dataset demonstrates the superiority of HGDiffuser over existing methods on generation efficiency.

